\documentclass[letterpaper, 10 pt, conference]{ieeeconf}  

\IEEEoverridecommandlockouts                              

\usepackage{graphics} 
\usepackage{epsfig} 
\usepackage{mathptmx} 
\usepackage{times} 
\usepackage{amsmath} 
\usepackage{amssymb}  
\usepackage{caption}
\usepackage{subcaption}
\usepackage{float}
\usepackage{placeins}
\usepackage{textcomp}
\usepackage{xspace}
\usepackage{xcolor}
\usepackage{booktabs}
\usepackage{multirow}
\usepackage{hyperref}
\let\labelindent\relax 
\usepackage[inline]{enumitem}
\usepackage{url}
\usepackage{microtype}

\newcommand{\mytilde}{\raise.17ex\hbox{$\scriptstyle\mathtt{\sim}$}}
\title{\LARGE \bf
MIDGARD: A Simulation Platform for Autonomous Ground Robot Navigation in Unstructured Environments
}

\author{Giuseppe Vecchio$^{1}$, Simone Palazzo$^{1}$, Dario C. Guastella$^{1}$, Ignacio Carlucho$^{2}$,\\
Stefano V. Albrecht$^{2}$, Giovanni Muscato$^{1}$ and Concetto Spampinato$^{1}$
\thanks{
$^{1}$Department of Electrical, Electronic and Computer Engineering, University of Catania, Italy.\endgraf
$^{2}$School of Informatics, University of Edinburgh, UK.\endgraf    
Corresponding author: \tt\small giuseppe.vecchio@phd.unict.it}}

\let\oldtwocolumn\twocolumn
\renewcommand\twocolumn[1][]{%
    \oldtwocolumn[{#1}{
    \begin{center}
           \includegraphics[width=\textwidth]{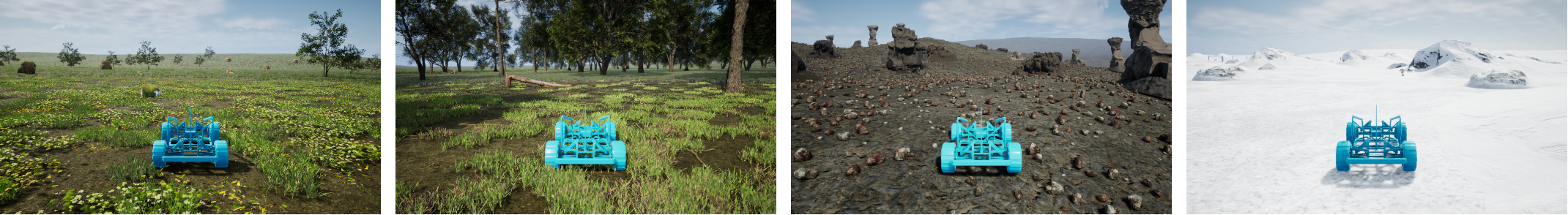}
           \captionof{figure}{Third-person captures of the four navigation environments available in MIDGARD.}
           \label{fig:thumbnail}
        \end{center}
    }]
}

\begin{document}

\maketitle

\thispagestyle{empty}
\pagestyle{empty}

\begin{abstract}
We present MIDGARD, an open-source simulation platform for autonomous robot navigation in outdoor unstructured environments. 
MIDGARD is designed to enable the training of autonomous agents (e.g., unmanned ground vehicles) in photorealistic 3D environments, and to support the generalization skills of learning-based agents through the variability in training scenarios.
MIDGARD's main features include a configurable, extensible, and difficulty-driven procedural landscape generation pipeline, with fast and photorealistic scene rendering based on Unreal Engine.
Additionally, MIDGARD has built-in support for OpenAI Gym, a programming interface for feature extension (e.g., integrating new types of sensors, customizing exposing internal simulation variables), and a variety of simulated agent sensors (e.g., RGB, depth and instance/semantic segmentation).
We evaluate MIDGARD's capabilities as a benchmarking tool for robot navigation utilizing a set of state-of-the-art reinforcement learning algorithms. The results demonstrate MIDGARD's suitability as a simulation and training environment, as well as the effectiveness of our procedural generation approach in controlling scene difficulty, which directly reflects on accuracy metrics.
MIDGARD build, source code and documentation are available at {\tt\small \url{https://midgardsim.org/}}. 
\end{abstract}

\section{Introduction}
\label{sec:introduction}

\begin{figure*}
    \centering
    \includegraphics[width=1\textwidth]{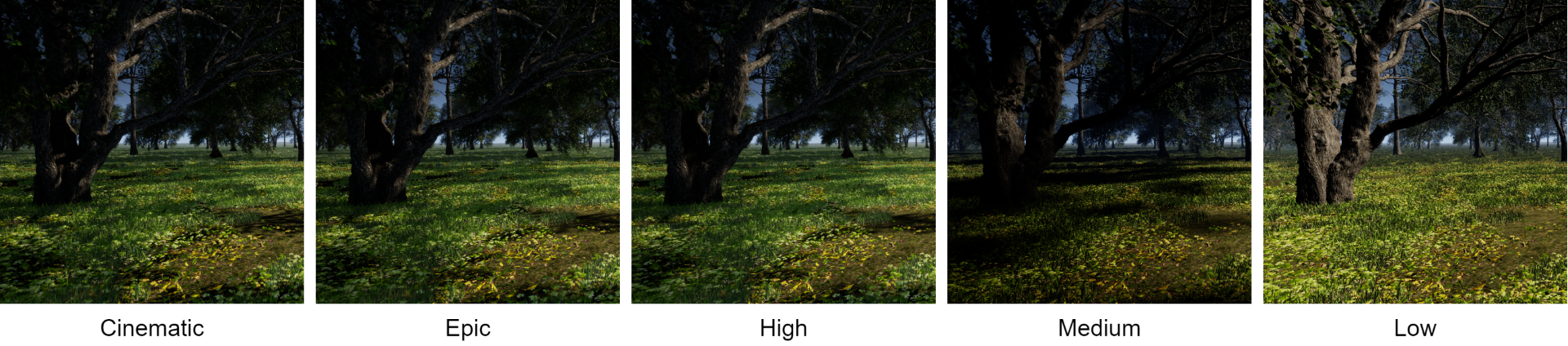}
    \caption{A comparison of the different rendering qualities available in MIDGARD. The difference becomes particularly evident in shadows and lighting quality when transitioning from \emph{``High''} to \emph{``Medium''} and from \emph{``Medium''} to \emph{``Low''}.}
    \label{fig:quality_comparison}
\end{figure*}

Autonomous ground robot navigation is still a major challenge in machine learning and robotics, especially in cases of outdoor unstructured and extreme environments, such as planetary exploration~\cite{Pflueger2019}, forest inventory~\cite{LostintheWoods}, search and rescue tasks~\cite{UMA-SARDataset} or precision agriculture~\cite{robotics9040097}. 
Indeed, while learning-based control models have been successfully applied for navigation in structured environments~\cite{CarluchoAUV,VGFNet,LidarEndToEnd}, their application to variable and highly-cluttered real-world scenes is far more complex~\cite{ReviewGuastella,borges22:fr}. This is mainly due to the  difficulty of collecting large amounts of data over a multitude of variable conditions and scenes. 
This makes real-world training and validation of learning models infeasible, due to high operative costs, risks of vehicle damage, slow training times, and limited scene variability, thus highlighting the need for simulation environments.

Simulation strategies have been adopted since the early days of research on autonomous navigation~\cite{pomerleau1998autonomous}. However, while many simulators exist for indoor navigation, e.g., HABITAT~\cite{savva2019habitat}, AI2-THOR~\cite{kolve2017ai2}, and House3D~\cite{wu2018building}, there is a general lack of simulators for outdoor unstructured environments. Most outdoor simulators, such as CARLA~\cite{dosovitskiy2017carla}, focus on urban scenarios. CARLA, in particular, is able to model the complexity of urban scenes  with a high-degree of realism; however, besides focusing on structured urban environments, CARLA does not support automatic scene generation, limiting the variability offered to users. Other existing simulators modelling unstructured environments are either not suitable for real-time environment interaction~\cite{muller2021photorealistic} or have been mainly designed for aerial vehicles~\cite{song2020flightmare}.

Motivated by the lack of available simulation platforms for navigation in unstructured environments, we introduce MIDGARD \footnote{``Kingdom of mankind``, i.e. Earth, in Norse mythology.},
an open-source, freely available simulation platform specifically designed for outdoor navigation in cluttered and unstructured environments. MIDGARD comes with an (extensible) set of pre-built environments,
it supports procedural generation with configurable scene difficulty, and provides a programming interface for under-the-hood customization (e.g., adding new types of sensors or accessing internal simulation parameters).
Compared to navigation in urban environments, where scene variability is somewhat limited and mostly given by the dynamics of pedestrians and vehicles, variability in unstructured environments is given by the scene itself. This makes the procedural generation of scenes a key element in the tackled task to encourage the agent to generalize over changing settings.
MIDGARD is based on Unreal Engine (UE)~\cite{unrealengine} and takes advantage of state-of-the-art technologies in rendering and simulation to provide a photorealistic navigation environment, which is a key element for transferring knowledge from the simulated environment to the real world. Finally, MIDGARD provides a Python interface compatible with OpenAI Gym~\cite{1606.01540}, enabling fast and reliable communication between agents and environment to support the training of machine learning models. 

To assess MIDGARD's capabilities as a benchmarking platform, we evaluate state-of-the-art reinforcement learning algorithms for autonomous navigation in unstructured environments under different simulation regimes. 
Our experiments show that MIDGARD provides a suitable and challenging simulation and training environment for reinforcement learning approaches; indeed, we demonstrate that the degree of success of navigation learning reflects the user-configurable difficulty of the environments. This confirms the validity of our procedural generation approach, which enables to investigate model behaviour under changing scene features.
\section{Related Work}

Simulation platforms have been proposed as a suitable alternative to training in the physical world for many tasks that require direct interaction with the environment~\cite{gaidon2016virtual,haltakov2013framework,richter2017playing,richter2016playing,skinner2016high,xia2018gibson}. 
In recent years, high-quality simulation engines have been presented: some of them focus on structured environments setups~\cite{savva2019habitat,dosovitskiy2017carla,shah2018airsim}, while others address unstructured environments~\cite{song2020flightmare}. Among the latter, the one closest to MIDGARD is CARLA~\cite{dosovitskiy2017carla}, a simulation engine for open urban driving based on Unreal. 
CARLA employs pre-built static urban environments, requiring manual crafting of new 3D scenes to increase variability, whereas training deep reinforcement learning models are known to require large and diverse training data~\cite{team2021open}.

In 2017, Shah et al. released AirSim~\cite{shah2018airsim}, a simulator for drones that also provides some support to cars and other vehicles. AirSim is open source, also built on Unreal, cross-platform, and supports software-in-the-loop simulation. It is developed as an Unreal plugin that exposes RPC APIs to interact with the vehicle in the simulation programmatically.
As such, it does not include any pre-built scene, thus requiring manual crafting, resulting in additional setup costs and limited training scenario variability.

HABITAT~\cite{savva2019habitat} makes a significant step forward to support ``training in simulation'' of artificial intelligence (AI) agents by adding flexibility to agent configuration, scene rendering, and control of the environment state. HABITAT focuses on embodied AI training in indoor, small-sized scenes and, thus, is unsuitable to outdoor navigation. Additionally, HABITAT lacks some features such as lighting control, and weather effects, which can be useful to extend scene variability.
Recently, M{\"u}ller et al. introduced OAISYS~\cite{muller2021photorealistic}, a photorealistic terrain simulation pipeline for unstructured outdoor environments, built on top of Blender~\cite{blender}. Like MIDGARD, OAISYS targets unstructured environments, but lacks real-time agent interaction, precluding the possibility to use it as a training framework for autonomous agents.

\section{MIDGARD Simulator}
\label{sec:simulator}

\begin{figure}[ht]
    \centering
    \includegraphics[width=1\linewidth]{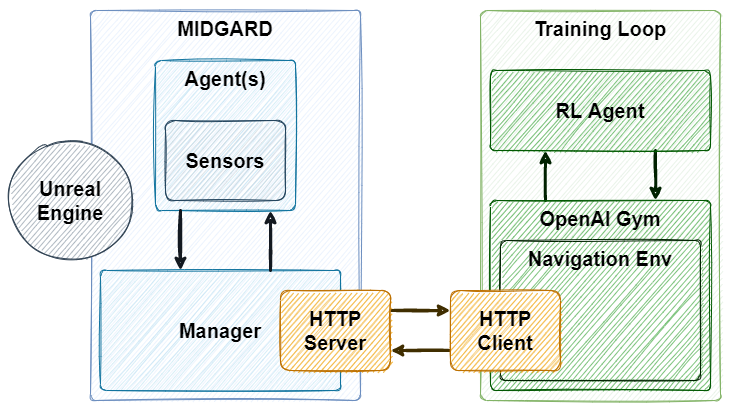}
    \caption{Overview of MIDGARD's architecture.}
    \label{fig:framework}
\end{figure}

MIDGARD is designed according to flexibility and performance principles. It provides a programming interface that allows users to configure and extend the core engine, the procedural scene generation, and a wide suite of ready-to-use sensors; additionally, it exposes internal state variables as training signals or for reward computation (e.g., distance from target and collision events).
A representation of MIDGARD architecture is presented in Fig.~\ref{fig:framework}.

\subsection{Engine features and configuration}

MIDGARD provides a set of general configuration options that are applied to all scenarios and affect the overall simulation. Rendering properties can be configured in terms of image resolution (from 128$\times$128 to 1024$\times$1024), image quality (from \emph{``Low''} to \emph{``Cinematic''}, as shown in Fig.~\ref{fig:quality_comparison}), and number of cameras mounted on the agent. Cameras can then be configured to capture specific types of signals, e.g. RGB or depth. These options directly affect the performance of the simulator, as discussed in detail in Sect.~\ref{sec:sim_perf}.

When configuring the environment, it is possible to set the dimensions of the simulated map (in square meters) and to choose between two types of virtual agent, with discrete or continuous controls, as described in Sect.~\ref{sec:agent}.
Finally, the simulation speed (i.e., the amount of time simulated by the engine at each step) can also be configured by the user.

\subsection{Procedural generation}

One of MIDGARD's main features is its capability to procedurally generate varying and dynamic simulated scenes. This feature is critical in the development of navigation algorithms able to perform well in the real world and that need to be robust to variations in the working environment (e.g., lighting and obstacle layout).
The scene generation process in MIDGARD represents the main advance with respect to existing simulators, as it is fully procedural and requires no human intervention other than an initial setup stage.

Procedural generation is based on the concept of \emph{scene type} (e.g., ``volcanic field''), formally defined by a \emph{scene descriptor}, which contains the specifications for generating \emph{instances} of that scene type. In practice, a scene descriptor consists of a \emph{base map}, defining a minimal characterization of the scene (for instance, in the case of ``volcanic field'', the base map may consist of an irregular surface with rocky terrain) and a set of \emph{world objects}, which can be placed all over the base map to generate a new random scene. The positioning of world objects is carried out by defining a \emph{grid} over the base map, and determining whether an object should be placed within each cell of the grid.

Formally, scene descriptors define the following attributes:
1) the base map for the scene type;
2) minimum and maximum grid cell size, in meters ($C_\text{min}$, $C_\text{max}$);
3) a set of placeable world objects.
In turn, each world object is defined by its 3D polygonal mesh and a set of attributes used at instantiation:
1) minimum and maximum size of the object;
2) maximum angle of inclination of the object with respect to the ground normal direction;
3) maximum ground slope: if the ground slope of a grid cell is higher than this value, the object cannot be placed.
These attributes allow scene creators to control the variability in object appearance and positioning, as well as the constraints that a realistic scene layout must satisfy (e.g., a tree should not be placed on a steep descent, nor it should be inclined by more than 20\textdegree-30\textdegree~on flat ground).

While new scene descriptors can be easily created with some basic knowledge of Unreal Engine, MIDGARD provides four pre-built different types of scenes for procedural generation, as shown in Fig.~\ref{fig:thumbnail}: 
\begin{itemize}[leftmargin=*]
\item \emph{Meadow}: Easy-medium difficulty navigation scene with obstacles like small or medium rocks and bushes. Slightly uneven surface covered in grass, which reduces obstacle visibility. It also features a large, non-traversable lake.
\item \emph{Forest}: Medium-high difficulty scene with dense obstacles like trees, rocks, broken branches, puddles and bushes. The surface is slightly uneven and covered in grass, woody debris, and leaves.
\item \emph{Volcanic field}: Medium-high difficulty scene with dense obstacles --- mostly rocks. It has an overall flat color, making low obstacles hard to distinguish from traversable surfaces. The surface is highly uneven and rocky.
\item \emph{Arctic glacier}: High difficulty scene, with sparse obstacles, such as snow-covered rocks and ice planes. The flat color of icy surfaces makes objects hard to distinguish.
\end{itemize}

At generation time, the environment receives a set of parameters by the Gym client, including:
1) difficulty level $D$, a parameter between 0 and 1 controlling obstacle density;
2) grid resolution $G_\text{res}$, a parameter between 0 and 1 that linearly interpolates cell size between $C_\text{min}$ and $C_\text{max}$;
3) for the default point-goal navigation task (discussed in Sect.~\ref{sec:results}), distance (in meters) between the agent's starting location and its target.

\begin{figure}[t]
    \centering
    \raisebox{-0.5\height}{\includegraphics[width=1\linewidth]{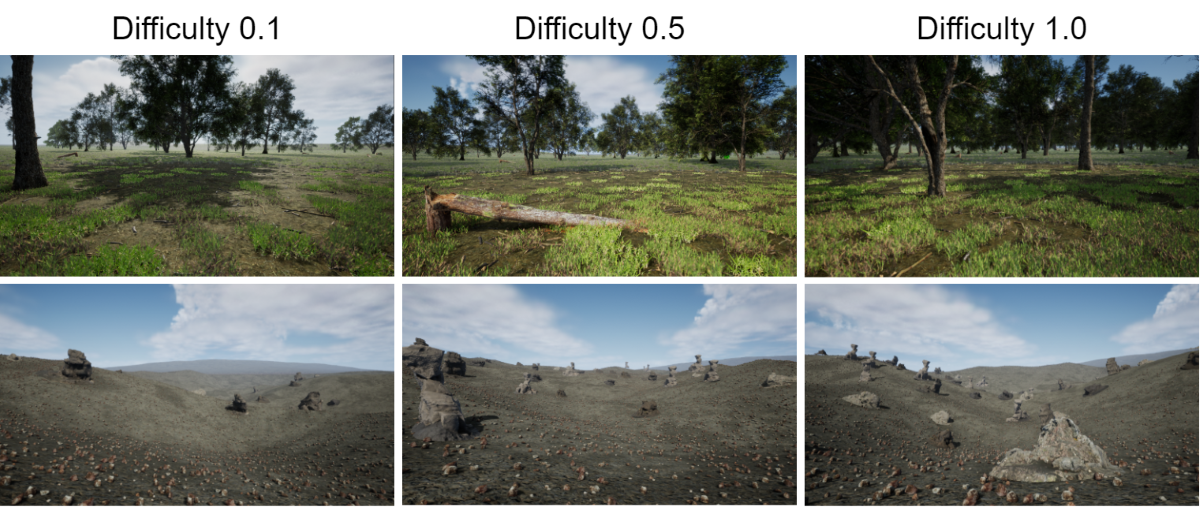}}
    \caption{Impact of the difficulty level on the obstacle density in the \emph{Forest} (top) and \emph{Volcanic} (bottom) scenes.}
    \label{fig:diff_comp}
\end{figure}

The map is divided into a grid with cells of size $C$, computed as:
\begin{equation}
    \centering
    C = C_\text{min} + G_\text{res}(C_\text{max}-C_\text{min}).
    \label{eq:cell_size}
\end{equation}
Then, for each cell, a world object is placed at a random location within the cell, with probability equal to the difficulty level $D$, as long as the ground slope at that location is within the limit defined in the object's descriptor. The type of object, its size and its inclination are set consistently with the specifications in the scene descriptor. Fig.~\ref{fig:diff_comp} provides examples of the effect of the difficulty level $D$ on the generation process.

\subsection{Virtual agent and sensor suite}
\label{sec:agent}

The virtual agent is an instance of an autonomous robot in the simulated scene, defined by two component modules: the \emph{perception module} and the \emph{control suite}.

\begin{figure}[t]
    \centering
    \includegraphics[width=1\linewidth]{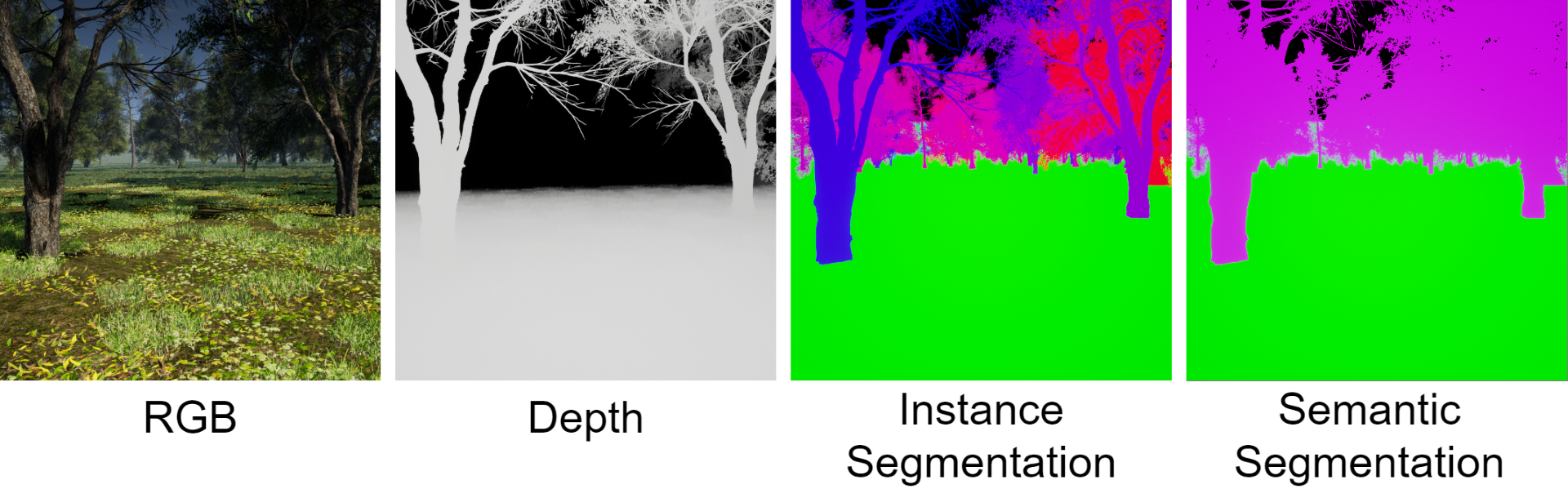}
    \caption{An example of the same view as perceived by the different camera sensors.}
    \label{fig:pixel_sensors}
\end{figure}

The MIDGARD perception module provides a full set of sensors designed for navigation. This module includes two types of sensors: 
\begin{enumerate}[leftmargin=*]
    \item Vision sensors in the form of on-board cameras that capture raw RGB pixel data, depth, and instance/semantic segmentation of the scene. Additionally, cameras have a set of parameters to control the field of view, the relative pose with respect to the vehicle, and the capture resolution. The semantic segmentation sensor provides segmentation areas categorized as ground, bush, tree, rock, water, debris, artificial.
An example of the view perceived through the different camera sensors is presented in Fig.~\ref{fig:pixel_sensors}.
\item  Low-level sensors that provide vector information of the agent state measurements and include: GPS-like vehicle location and orientation in the map reference frame, navigation target location, inertial measurements (attitude, speed, and acceleration), and collision detection.
\end{enumerate}

As for the control suite, MIDGARD includes two types of 4-wheel vehicles, which support different types of control actions:
1) \emph{discrete}: controlled by a set of discrete actions that include brake, forward, backward, turn left, turn right;
2) \emph{continuous}: controlled through a continuous action set: brake, linear speed, and angular speed.
Actions in the environment are performed until a new action is received, providing asynchronicity between the agent and the environment, similarly to what would happen in the real world.

\subsection{Simulator performance}
\label{sec:sim_perf}

While the entire simulation and rendering mechanism is efficiently handled by Unreal Engine, several key components of MIDGARD have been built from scratch, including image capture and transmission, procedural scene generation and obstacle positioning, and virtual agent implementation. These components have been written in C++ and optimized to provide a high frame rate and low response times.

For clarity and transparency in analyzing MIDGARD's performance, we distinguish between the \emph{simulation step rate} and the \emph{frame capture rate}. The former is related to the speed at which the simulator internally advances and renders the scene. For example, at \emph{``High''} image quality (see Fig.~\ref{fig:quality_comparison}) MIDGARD runs about 120 simulation steps per second. Frame capture rate, instead, is the frequency at which images are captured from the simulation. This operation includes an overhead due to data sampling and transmission, causing the frame capture rate to be lower than the simulation step rate. For practical purposes, users are more interested in the frame capture rate, which is reported in Tab.~\ref{tab:capture_fps} for different combinations of image quality, resolution, and number of cameras. Using a reasonable setup, i.e., \emph{``High''} image quality, 512$\times$512 pixel resolution and one camera, MIDGARD is able to provide 60 frames per second, which is sufficient for most practical applications.

\begin{table}[!htbp]
\centering
\begin{tabular}{cccccc}
\toprule
& & \multicolumn{4}{c}{\textbf{Resolution}} \\
\cmidrule{3-6}
\textbf{Quality} & \textbf{Cameras} & \textbf{128} & \textbf{256} & \textbf{512} & \textbf{1024} \\
\midrule
{\multirow{2}{*}{\textbf{Cinematic}}}
& \textbf{1} & 61 & 58 & 54 & 35 \\
& \textbf{2} & 48 & 44 & 38 & 18 \\
\midrule
{\multirow{2}{*}{\textbf{Epic}}}
& \textbf{1} & 64 & 62 & 58 & 37 \\
& \textbf{2} & 52 & 48 & 42 & 20 \\
\midrule
{\multirow{2}{*}{\textbf{High}}}
& \textbf{1} & 76 & 72 & 60 & 40 \\
& \textbf{2} & 58 & 54 & 45 & 22 \\
\midrule
{\multirow{2}{*}{\textbf{Medium}}}
& \textbf{1} & 78 & 75 & 66 & 42 \\
& \textbf{2} & 66 & 60 & 50 & 24 \\
\midrule
{\multirow{2}{*}{\textbf{Low}}}
& \textbf{1} & 110 & 92 & 70 & 45 \\
& \textbf{2} & 70 &  64 & 52 & 27 \\
\bottomrule
\end{tabular}
\caption{Frame capture rate for different combinations of image quality, number of cameras, and resolution. Computed on a Windows 11 machine with a Ryzen 5950X, Nvidia RTX 3090, 64 GB RAM and PCIe SSD.}
\label{tab:capture_fps}
\end{table} 

\begin{table*}[ht!]
\centering
\begin{tabular}{ll|cccc}
\toprule
& \textbf{Feature}  & \textbf{AirSim} & \textbf{Habitat} & \textbf{CARLA} & \textbf{MIDGARD} \\
\midrule
{\multirow{3}{*}{\rotatebox[origin=c]{90}{\textbf{Perform.~}}}}
& Simulation step rate & 120 & 8,000 & 120 & 120 \\
& Frame capture rate & Not specified & Not specified & 60 & 60 \\
& Scene creation time & Static & Static & Static & 1 sec \\
& Visual photorealism & Scene-dependent & Moderate & Moderate & High \\
\midrule
{\multirow{5}{*}{\rotatebox[origin=c]{90}{\textbf{~~~Flexibility}}}} 
& Scene type  & None provided & Indoor & Urban Outdoor & Unstructured Outdoor \\
& Assets import & Pre-build & Runtime & Pre-build & Pre-build \\
& Scene variability & None & None & None & At runtime (procedural) \\
& Scene difficulty & Intrinsic in scene & Intrinsic in scene & Intrinsic in scene & Controllable \\
\midrule
{\multirow{4}{*}{\rotatebox[origin=c]{90}{\textbf{~~~API}}}} 
& Gym environment & Not officially & Yes & Not officially & Yes\\
& Features extension & Not documented & None & Not documented & C++/Blueprint \\
& Expose new APIs & None & None & None & As HTTP endpoint \\
\bottomrule
\end{tabular}
\caption{Comparison between MIDGARD (using average settings) and other simulators.}
\label{tab:comparison}
\end{table*} 

\subsection{Comparing MIDGARD to existing simulators}

In this section, we compare MIDGARD to existing simulators in terms of performance, flexibility and available features.
An overview of the comparison between MIDGARD and  AirSim~\cite{shah2018airsim}, Habitat~\cite{savva2019habitat} and CARLA~\cite{dosovitskiy2017carla} is given in Tab.~\ref{tab:comparison}. Performance assessment is carried out in terms of:
\begin{enumerate}[leftmargin=*]
    \item \textbf{Simulation step rate}:
    Unreal-based platforms achieve about 120 FPS (frames per second), with average quality settings. Habitat reaches higher FPS forgoing dynamic and indirect illumination, complex material shading, weather effects, volumetric effects, etc.~\footnote{\url{https://aihabitat.org/docs/habitat-sim/lighting-setups.html}}.
    \item \textbf{Frame capture rate}: about 60 FPS in a single-camera setup for both CARLA and MIDGARD (no official data available for AirSim and Habitat).
    \item \textbf{Scene creation time}: MIDGARD's procedural generation takes about 1 second to produce a scene; the other simulators employ static, hand-crafted scenes.
    \item \textbf{Visual realism}: MIDGARD, CARLA, and AirSim (which, however, does not provide built-in scenes) leverage videogame technology to deliver photorealism, achieving better results than Habitat.
\end{enumerate}

Flexibility is evaluated in terms of: 
\begin{enumerate}[leftmargin=*]
    \item \textbf{Scene type}: each simulator tackles a specific setting, except AirSim which is general but does not provide built-in scenes.
    \item \textbf{Asset imports}: at build-time for Unreal-based simulators, i.e., MIDGARD, CARLA, and Airsim; Habitat provides runtime asset loading.
    \item \textbf{Scene variability}: only MIDGARD supports scene variations through procedural generation.
    \item \textbf{Scene difficulty}: MIDGARD supports scene difficulty variations, by controlling obstacle density in the scene instance.
\end{enumerate}

Regarding platform integration and extensibility:
\begin{enumerate}[leftmargin=*]
    \item \textbf{OpenAI Gym support}: built-in in MIDGARD and Habitat; unofficial repositories are available for AirSim and CARLA.
    \item \textbf{Feature extension}, i.e., integrating new functionalities, such as custom vehicles and sensors: supported in C++ and Blueprint by MIDGARD, and theoretically supported but not documented by AirSim and CARLA.
    \item \textbf{Exposing custom functions}: not officially supported by other simulators. MIDGARD can expose Blueprint functions as HTTP endpoints that can be called from any external application.
\end{enumerate}

\section{MIDGARD for RL-based Autonomous Navigation}
\label{sec:results}
In this section, we describe the employment of  MIDGARD in training two state-of-the-art reinforcement learning methods, namely, PPO~\cite{schulman2017proximal} and SAC~\cite{haarnoja2018soft}, at a navigation task.

\subsection{Task definition}
We tackle the task of point-goal navigation, where the agent has to go from point \textit{A} to point \textit{B}, both chosen randomly under the constraint that they have to be in traversable and object-free locations. The location of point \emph{B} is assumed to be known by the virtual agent.
Visual information is acquired through a camera sensor positioned one meter above the ground and tilted towards the terrain at an angle of 10\textdegree, with a field of view of 90\textdegree. Images are captured at a resolution of 256$\times$256 pixels. GPS information regarding agent and target positions is provided to the agent as a vector with the current distance to the target (normalized with respect to the map size) and heading angle. The agent is also provided with the difference between the distances from the target at consecutive steps, as a simple way to embed temporal information of its navigation trajectory.
The environment provides the agent with end-of-episode signals, notifying when it either reaches the target, collides with an obstacle, or crosses a non-traversable region. To prevent stalling behavior by the agent, episodes last up to 500 time steps.

\subsection{Policy network architecture}

\begin{figure}[t]
    \centering
    \includegraphics[width=1\linewidth]{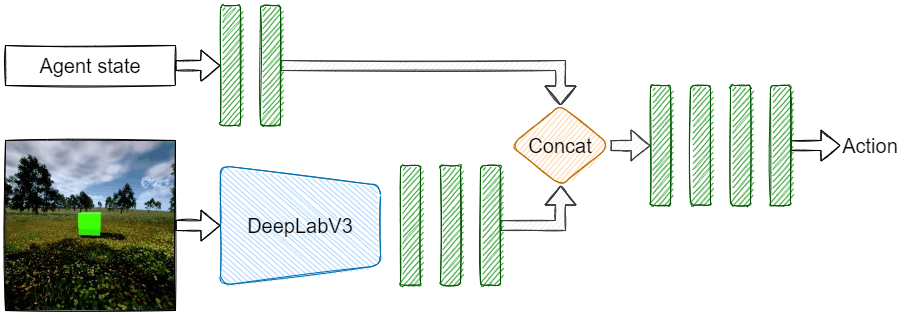}
    \caption{Overview of the policy network architecture. Two separate branches process visual data and agent state, before concatenating the extracted features and predicting the next action to perform.}
    \label{fig:network}
\end{figure}

As a visual feature extractor, we employ a DeepLabV3~\cite{chen2017rethinking} encoder, pretrained on MS-COCO~\cite{lin2014microsoft}, as available in the PyTorch library\footnote{\url{https://pytorch.org/hub/}}. The produced 8$\times$8$\times$960 embedding is spatially squeezed through adaptive max pooling, and fed to a cascade of three fully-connected layers (of size 128, 64, 64). Another network branch processes agent state features (distance, angle and distance difference) through two fully-connected layers of size 64. Visual and state feature are then concatenated and fed to a cascade of four fully-connected layers (of size 128, 64, 64, and 5, for a discrete-action agent) to predict the action to perform. All fully-connected layers apply ReLU activations. An overview of the network architecture is presented in Fig.~\ref{fig:network}.

\subsection{Training details}

The reward function for both PPO and SAC is designed to incentivize the agent to reach the target location with a limited amount of steps. It consists of a positive reward when successfully reaching the target (+10), a negative reward when a collision is detected or when non-traversable surfaces are crossed (-10), and a small negative signal for each step (-0.05) or when the agent gets further from the target (-0.1).  

Training is carried out on standardized RGB data and agent information, for 20,000 episodes.
Reward discount factor is set to 0.99. The simulation step rate is set to 10 (i.e., the engine simulates 100 ms at each step). Episodes are stopped after 500 steps if the agent does not reach the target nor collide with obstacles. The PPO agent is trained every 2,500 steps and each update consists of 80 epochs. We use an initial learning rate of 0.0003 for the actor and 0.001 for the critic. The SAC agent is trained every 4 steps and the network is updated, similarly to PPO, every 2,500 steps with a batch of 64 elements. We train two Q-functions, which was proven by Haarnoja et al.~\cite{haarnoja2018soft} to speed up the training, especially in complex tasks. The learning rate is initially set to 0.0003 for both critics, the policy network and the feature extractor.

\subsection{Results}

Performance evaluation of PPO~\cite{schulman2017proximal} and a discrete implementation of SAC~\cite{haarnoja2018soft} on MIDGARD's default scenes is measured in terms of average reward, success rate and average number of steps per episode. We choose \emph{Forest} and \emph{Volcanic field} as benchmark scenes, to showcase the performances of state-of-the-art algorithms in two medium difficulty environments. Experiments are conducted at different difficulty levels, to explore agent behaviour in increasingly cluttered settings. Additionally, for each difficulty level, we evaluate performances at a distance of 10 and 20 meters from the target. Tab.~\ref{tab:reward},~\ref{tab:success} and~\ref{tab:steps} report these metrics for different settings of target distance and scene difficulty. Both algorithms are able to solve the tackled task with a success rate above 50\% in setting with difficulty level of 0.1 and distance to target of 10 meters (SAC reaches up to 62\%). Performance starts to quickly degrade, in both scenes, when increasing the difficulty level or the distance to target. SAC shows a slight advantage over PPO in most settings, with the gap being more prominent at higher difficulty levels and greater starting distance from the target.

Overall, these results confirm that MIDGARD provides a challenging benchmark to state-of-the-art reinforcement learning methods. Interestingly, they also validate the effectiveness of our approach for configurable scene difficulty, which directly reflects on the performance of the navigation algorithms. 

\begin{table}[t]
\centering
\begin{tabular}{lccccc}
\toprule
& & & \multicolumn{3}{c}{\textbf{Difficulty}} \\
\cmidrule{4-6}
\textbf{Scene} & \textbf{Agent} & \textbf{Distance} & \textbf{0.1} & \textbf{0.5} & \textbf{1.0} \\
\midrule
\multirow{4}{*}{\emph{Forest}}   & \multirow{2}{*}{PPO} & 10 m & -3.56 & -6.41 & -16.46 \\
                                 &                      & 20 m & -13.31 & -16.02 & -21.92 \\
                                 \cmidrule{2-6}
                                 & \multirow{2}{*}{SAC} & 10 m & -2.78 & -4.48 & -7.85 \\
                                 &                      & 20 m & -6.45 & -10.59 & -14.29 \\
\midrule
\multirow{4}{*}{\emph{Volcanic}} & \multirow{2}{*}{PPO} & 10 m & -4.86 & -8.92 & -13.18 \\
                                 &                      & 20 m & -10.77 & -13.40 & -18.67 \\
                                 \cmidrule{2-6}
                                 & \multirow{2}{*}{SAC} & 10 m & -2,73 & -5,48 & -8,42 \\
                                 &                      & 20 m & -7,46 & -10,37 & -15,51 \\
\bottomrule
\end{tabular}
\caption{Average total reward per episode after training, at different difficulty levels and distances from target.}
\label{tab:reward}
\end{table}

\begin{table}[t]
\centering
\begin{tabular}{lccccc}
\toprule
& & & \multicolumn{3}{c}{\textbf{Difficulty}} \\
\cmidrule{4-6}
\textbf{Scene} & \textbf{Agent} & \textbf{Distance} & \textbf{0.1} & \textbf{0.5} & \textbf{1.0} \\
\midrule
\multirow{4}{*}{\emph{Forest}}   & \multirow{2}{*}{PPO} & 10 m & 54\% & 42\% & 23\% \\
                                 &                      & 20 m & 41\% & 34\% & 15\% \\
                                 \cmidrule{2-6}
                                 & \multirow{2}{*}{SAC} & 10 m & 62\% & 51\% & 37\% \\
                                 &                      & 20 m & 53\% & 39\% & 27\% \\
\midrule
\multirow{4}{*}{\emph{Volcanic}} & \multirow{2}{*}{PPO} & 10 m & 48\% & 37\% & 17\% \\
                                 &                      & 20 m & 38\% & 29\% & 15\% \\
                                 \cmidrule{2-6}
                                 & \multirow{2}{*}{SAC} & 10 m & 60\% & 49\% & 33\% \\
                                 &                      & 20 m & 48\% & 37\% & 22\% \\
\bottomrule
\end{tabular}
\caption{Success rate after training, at different difficulty levels and distances from target.}
\label{tab:success}
\end{table}

\begin{table}[t]
\centering
\begin{tabular}{lccccc}
\toprule
& & & \multicolumn{3}{c}{\textbf{Difficulty}} \\
\cmidrule{4-6}
\textbf{Scene} & \textbf{Agent} & \textbf{Distance} & \textbf{0.1} & \textbf{0.5} & \textbf{1.0} \\
\midrule
\multirow{4}{*}{\emph{Forest}}   & \multirow{2}{*}{PPO} & 10 m & 78.4 & 110.7 & 156.8 \\
                                 &                      & 20 m & 226.5 & 243.2 & 258.5 \\
                                 \cmidrule{2-6}
                                 & \multirow{2}{*}{SAC} & 10 m & 74.6 & 92.9 & 125.4 \\
                                 &                      & 20 m & 137.9 & 174.1 & 196.3 \\
\midrule
\multirow{4}{*}{\emph{Volcanic}} & \multirow{2}{*}{PPO} & 10 m & 94.2 & 142.0 & 159.1 \\
                                 &                      & 20 m & 174.5 & 188.7 & 237.8 \\
                                 \cmidrule{2-6}
                                 & \multirow{2}{*}{SAC} & 10 m & 69.1 & 108.4 & 126.7 \\
                                 &                      & 20 m & 144.3 & 165.5 & 199.4 \\
\bottomrule
\end{tabular}
\caption{Average episode length after training, at different difficulty levels and distances from target.}
\label{tab:steps}
\end{table}
\section{Conclusion}
\label{sec:conclusion}

In this work, we presented MIDGARD, an open-source photorealistic simulation environment, based on Unreal Engine, for supporting research in autonomous navigation in outdoor unstructured environments.

One of MIDGARD's main features is its novel procedural scene generation with configurable difficulty, which produces infinite scene variations to improve the generalization capabilities of learning agents.
Additionally, MIDGARD is equipped with an OpenAI Gym front-end, for an easy-to-use interface with machine learning models, and exposes a programming interface for extending its functionalities, in terms of available scenes, virtual agent sensors and access to internal state variables.
In our experiments, MIDGARD proved to be a challenging navigation benchmark for state-of-the-art reinforcement learning methods, which achieved a success rate lower than 65\%. However, these results demonstrate the validity of our approach for configurable difficulty, which directly reflects on the resulting success rate, and which could be employed to support more effective curriculum learning strategies.

MIDGARD's current limitations, to be addressed in our future work, concern performance, agent customization and simulation realism.
Multiple-agent spawning in the same scene will speed up and increase the variability of experience at training time, as well as to provide an execution environment for further navigation-related tasks, such as collaborative SLAM techniques~\cite{ZOU2019461} or coverage path-planning~\cite{perez16:case} in unstructured environments.
We also plan to support multiple robot models (e.g., 4-wheeled, multi-wheeled, tracked, etc.) and to improve the simulation of agent-terrain interaction, in order to obtain a behaviour closer to the real world.
Additionally, MIDGARD will also feature bi-directional communication with ROS (Robot Operating System), opening up to further solutions for the external control, monitoring and testing of the vehicle.
Finally, in order to solve the navigation task under sparse rewards and with limited experience, future developments on MIDGARD will include human-in-the-loop control, which can be leveraged by imitation learning algorithms~\cite{ILSruvey}.

\bibliographystyle{IEEEtran}
\bibliography{bibliography}

\end{document}